# Employing traditional machine learning algorithms for big data streams analysis: the case of object trajectory prediction

*Angelos Valsamis[1], Konstantinos Tserpes[2], Dimitrios Zissis[3], Dimosthenis Anagnostopoulos[2], Theodora Varvarigou[1]*

[1] Dept. of Electrical and Computer Engineering,
National Technical University of Athens
Heroon Polytechniou 9, Zografou, Greece
ang.valsamis@gmail.com, dora@telecom.ntua.gr

[2] Dept. of Informatics and Telematics
Harokopio University of Athens
Tavros, Greece
{tserpes,dimosthe}@hua.gr

[3] Dept. of Product and Systems Design Engineering
University of the Aegean
Ermoupolis, Syros, Greece
dzissis@aegean.gr

**Abstract**

In this paper, we model the trajectory of sea vessels and provide a service that predicts in near-real time the position of any given vessel in 4', 10', 20' and 40' time intervals. We explore the necessary tradeoffs between accuracy, performance and resource utilization are explored given the large volume and update rates of input data. We start with building models based on well-established machine learning algorithms using static datasets and multi-scan training approaches and identify the best candidate to be used in implementing a single-pass predictive approach, under real-time constraints. The results are measured in terms of accuracy and performance and are compared against the baseline kinematic equations. Results show that it is possible to efficiently model the trajectory of multiple vessels using a single model, which is trained and evaluated using an adequately large, static dataset, thus achieving a significant gain in terms of resource usage while not compromising accuracy.

# 1 Introduction

Building data analytics methods can assist in the task of modeling vessel behavior and responding to questions related to its future state; including predicting its position in the next few minutes. "Normal" vessels patterns can be used to identify potential deviations from these due to suspicious conditions. This kind of information can be of paramount importance as abnormal trajectories tend to hold early indicators of potential problems, which require immediate attention and need to be resolved at an early stage. As such, there are many applications that require real-time monitoring of abnormal sequential patterns over streaming trajectories. In this particular work we employ a proof-of-concept use case, regarding the prediction of trajectories of vessels at sea. This problem falls into the wider research area referred to as Trajectory Prediction (TP) and even considering past research, it remains a very challenging exercise [1] mainly due to the additional requirements of high speed, data stream processing within limited space cost [2] and time constraints.

Data streams differ from traditional finite data sets in that they are temporally ordered, fast changing, massive, and potentially infinite sequences of real time data that are generated from non-stationary distributions in dynamic environments. A stream-processing algorithm is evaluated according to the number of passes the algorithm must make over the data, the available memory and the execution time. In real-time analysis, queries should run continuously; logically issued once, but run forever [3]. The answer to a continuous query is produced over time, reflecting the data seen so far or seen in windows of more recent data. To discover knowledge or patterns from data streams, it is necessary to develop single-scan, on-line, multilevel, multidimensional stream processing and analysis methods [4]. Learning algorithms have to exhibit characteristics of incremental and adaptive learning incorporating concept of drift, forgetting outdated data and adapting to the most recent state of the nature to be relevant [5]. Typical methods used in this rationale are Random sampling, Histograms, Sliding windows, Multi-resolution model, Sketches and Randomized algorithms.

A generally acceptable assumption is that traditional methods of frequent item set mining, classification, and clustering tend to scan the data multiple times, making them inadequate for stream data [4]. On the contrary, when mining data streams the mechanisms operate on top of user-defined tradeoffs between –at least- accuracy, performance and resource usage. This implies that data stream algorithms often need to sacrifice accuracy in order to meet basic application requirements and frequently they resolve to probabilistic models.

This assumption was the premise for our previous work [6] in which we built a system capable of predicting multiple vessels behavior from streaming data under real-time constraints, employing single scan, fixed window algorithms and a toolkit called Massive Online Analysis (MOA) [7]. MOA includes a collection of machine learning algorithms and tools for evaluation and it takes advantage of current PC architectures in order to scale effectively and deliver timely results.

In this work, we evaluate this assumption, employing traditional, multi-scan, pre-trained models for implementing online predictive services that process data streams. The main goal is to investigate the tradeoff between accuracy, performance and resources and compare it to single-scan, big data frameworks. This work provides an analysis for researchers or practitioners attempting to select the most appropriate models for AIS or similar datasets (such as ADS-B aviation generated data). We experiment with multiple traditional machine learning algorithms and then test them in real-time response settings, using time-series data streams. An important notion is that the proposed tool must be generalizable to any application domain that requires predictive analytics, independently of the data characteristics. As such, the only element for constructing the predictive model is the data itself, i.e. no application characteristics are considered (e.g. geo-location). This infuses a new challenge in solving the problem, that is, a single model must be able to predict the location of any vessel.

We consider that the main contribution of this work is to show that a well-trained model that achieves high accuracy predictions may very well also meet the real-time response requirements of a predictive service without the need for retraining it and without sacrificing accuracy. This is a counter-intuitive statement as many works seek move away from traditional machine learning techniques for the sake of tailored made solutions that hurt accuracy in the altar of performance.

The document is structured in 5 sections. Section 1 is the current Introduction. Section 2 analyzes the related work in the fields of time-series, real-time analysis tools and traditional trajectory prediction models. Section 3 provides the details of the proposed approach and Section 4 explains the evaluation framework and results. Finally, Section 5 summarizes the major conclusions from this experiment.

## 2  Related Work

The related work can be separated in two major classes, depending on the angle from which the problem is investigated. From a data stream point of view, we sought for relevant solutions in the time-series, real time analysis domain, while at the same time we studied the traditional algorithms employed for the problem of trajectory prediction. In what follows we present our research on these two classes. To the best of our knowledge, there is no single work in the current literature comparing real time big data predictive analytics for trajectory prediction.

### 2.1  Time-series, Real-time Analysis Tools

Incoming data needs to be processed immediately as it becomes available, as its high velocity and volume makes it impossible to store in active storage (i.e., in a conventional database) and interact with it at a time of our choosing. There are two logical methods of process data streams:

- Stream summarization: Maintaining only a small amount of elements which characterize the data stream (and often the application we are tackling) and which requires much smaller space and time to be processed
- Process fixed length windows: Fixed portions of the data are loaded into a memory buffer and machine learning algorithms are applied only to that buffer, perhaps also considering past performance for optimization purposes

In terms of architecture, in both the above cases, the stream processor maintains a working storage of manageable size; "manageable" being defined by the application requirements, in terms of time constraints and accuracy of expected result as well as available resources.

However, in this work we are interested in time-series forecasting, i.e. discovering the mathematical formula which will approximately generate the historical patterns in a time series [8]. Our goal is to build an accurate model for predicting an object's trajectory; thus learning the object's kinematic equations. This makes the fixed length window processing more relevant and thus, the remainder of this Section focuses on that.

The common approaches for time-series forecasting were mainly based on linear statistics with a class of models called ARIMA (Auto-Regressive Integrated Moving Average) being a popular example [9]. Numerous variations of ARIMA have been presented in the literature for forecasting models using data streams including [10], [11] and [12]. However, the recent up taking of distributed computing techniques, mainly due to the emergence of cloud computing, led industry and scientists alike to turn their interest to traditional machine learning algorithms applied in a distributed fashion. This ensured that all the benefits from well established mechanisms would be maintained even when applied to a different context. Furthermore, it allowed data analysts to remain in their "comfort zone" when dealing with data streams since they were called to use traditional machine learning tools which they were accustomed with already.

As such, general frameworks have been developed for supporting massive stream versions of typical machine learning algorithms. For instance, Domingos et al have developed one such framework [13] for data stream mining using decision tree induction, Bayesian network learning, k-means clustering and the EM algorithm for mixtures of Gaussians has been developed.

More recently, similar software packages for forecasting models have appeared such as Massive Online Analysis (MOA) [7][14] an open source framework for real-time stream analytics, Rapid-Miner [15][16] a data mining system with plug-in for stream processing, MEKA [17][18], a multi-label extension to the popular WEKA library for machine learning, etc.

## 2.2 Trajectory Prediction Models

Regarding models for forecasting moving models and maneuvers, used in recent literature, they can be divided into three categories considering their dimensional space: 1-D, 2-D, and 3-D models. The 3-D models are commonly used in air navigation systems, and 1-D and 2-D models are commonly used in land and maritime navigation systems [19]. When a single model cannot capture the required behavior of a target, a multiple model approach has also been proposed. In [20] Ristic et al, use historic AIS vessel motion pattern data to predict a vessel's motion based on the Gaussian sum tracking filter. Perera et al [19] implement an extended Kalman filter (EKF) for the estimation of vessel states and for the prediction of vessel trajectories. Vanneschi et al [21], propose a novel computational intelligence framework, based on genetic programming, so as to predict the position of vessels, based on information related to the vessels past positions in a specific time interval. Duh and Lin developed standard Kalman filter with a self-constructing neural fuzzy inference network

(KF-SONFIN) algorithm for target tracking, capable of overcoming low convergence problems and large network size of Artificial Neural Network (ANN)[22].

Focusing on ANNs, their unique characteristics– adaptability, nonlinearity, arbitrary function mapping ability – make them quite suitable and useful for forecasting tasks. As such, they have found multiple applications in traffic movement modeling such as the forecasting of the traffic flow at the Suez canal [23]. In [24] García et al perform an extended analysis of published prediction models used to model transport systems and in particular in for predictions in port areas. The comparisons of ANNs with other forecasting models indicate better results for ANNs. Recently, Zissis et al, [25] implemented a web based ANN capable of predicting a vessel's positions in near real-time requirements which operated on static vessel data collected in a short past time window.

Based on this analysis, in this work, we sought for a middle ground solution that combines the benefits of real time spatiotemporal mining systems and the most appropriate trajectory prediction algorithms. This setup was able to preserve the time and resource constraints while yielding accurate results. Those results were further improved by optimizing several parameters. The details are explained below.

# 3 Approach

As already mentioned, the key challenge that this work faced, revolved around the real time prediction of the trajectories of sea-vessels using real data streams reporting their position, heading, ID and some other metadata.

This research is motivated by the fact that providing valid and accurate estimations of future vessel positions, comprise valuable knowledge for decision making processes for ship owners, insurance companies, coastguards, port management, etc. The forecasting model is established upon the premises that past behavior of a vessel can assist in inferring knowledge about its future position. Since we are referring to data streams, it is crucial to clarify that the definition of past behavior is subtle and it largely depends on the time and computational resource constraints. As shown in [26], when large amounts of historic movement records are available, and multiple scans on the data can be tolerated, then the results can be remarkably accurate.

## 3.1 Baseline work

In our previous work [6] we employed a fixed time window stream-processing model and applied a number of semi-supervised learning methods. The key concept is that for each batch fed into the algorithm, the prediction was measured against the actual position, which was then reported in the immediate future, employing an iterative model refinement process. Thus, the prediction algorithm performance was such that the result came at least one time unit before the record bearing the predicted position.

Each algorithm made a prediction using the fixed window batch and evaluated itself completing a training step against the actual position. All algorithms suffered from the cold start problem, especially because there was no distinction between the vessels semantics, i.e. each vessel record was treated as a record equally contributing to the model. This implies that there was no separation of models between vessel types, geographic area or any other distinguishing parameter. Even though this would greatly improve the accuracy of the result as reported in [26] but also based on mere intuition, such approach was not followed because of the computational burden that would pose to the underlying infrastructure. As such, we created our single pass, real-time responding model in MOA, employing a simple perceptron, conveniently leveraging on the system's architecture to scale effectively and deliver timely results.

## 3.2 Assumptions and approach

The current work investigates the limitations and strengths of using traditional machine learning algorithms for predictive services in big data stream problems. As mentioned earlier, the purpose is to create a single model that will effectively predict vessel locations and integrate that in a system that will respond to queries in real time. The queries essentially involve the provision of the present location and possibly other information by the end user to which the system must reply by a location prediction in the next 2', 4', 20' and 40'. The reply must provided in real time, i.e. in less than 0.1 s. The user must feel at all times that the system is reacting instantaneously, a commonly accepted user experience metric defined by Miller [27].

The approach followed for implementing such a service involves the following steps:

- Rank a range of machine learning algorithms based on their ability to predict accurately trajectories in a fixed, large volume dataset, employing multiple passes over the dataset.
- Use the best performing trained model so as to implement a service that will use streams of data in order to provide a future (2', 4', 20', 40') prediction that is both accurate but also generated as soon as it is requested, thus providing a near-real time service.
- Compare the performance and accuracy of the selected algorithm and its big data implementation to the respective of a service implementing a single service simply implementing the kinematic equations as well as a fixed time window stream-processing model as described in [6].

Our motivation behind using traditional machine learning algorithms, and not fully utilizing the nature of time series data, is the natural scarcity of our dataset. Take for instance the case of the 2' forecasting interval. In that case our dataset drops from 1.315.474 AIS entries to 569.680 instances. We effectively lose more than half of our data because there is no symmetry in the frequency of the AIS entries, i.e. half of our ship entries do not have a pair with a 2' difference. Considering that we will need to split the remaining instances in order to be able to evaluate our model, the entries available for training are 434.606 (consider a 75/25 split), which is one third of our initial dataset.

If we choose to handle the remaining dataset as a time series we are going to lose even more of our training power. That is true, because we would rather perform listwise deletion rather than imputation or maximum likelihood methods. The reason behind this is that we focus on maximizing accuracy in the evaluation of our model, which can be penalized by substituting missing values. Furthermore, in the real time scenario, performing those methods can be time and space consuming. We argue that preserving a memoryless state can help in the scalability of our service and the generalization of our model.

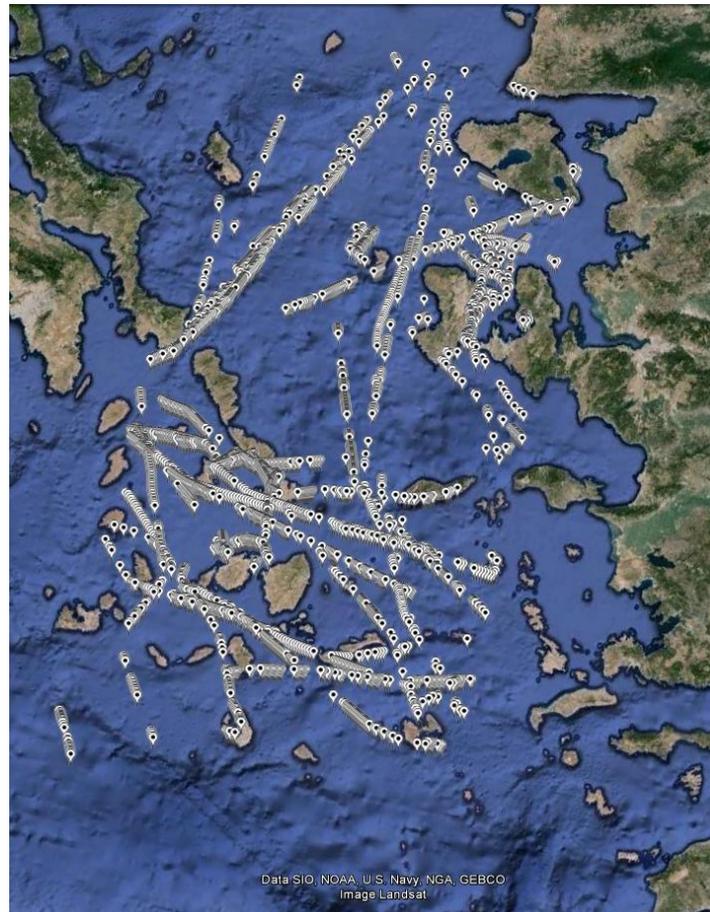

Figure 1 - Sample from our dataset involving 1,315,474 AIS entries, regarding 2874 unique vessels in the Aegean Sea

The data used for this study is provided by MarineTraffic.com and is based on AIS. AIS transmissions can be defined as spatial time series [28], describing the movements of vessels across geographic regions. An AIS message contains the vessel's Maritime Mobile Service Identity (MMSI) – a unique nine digit identification number; Navigation status; Rate of turn; Speed over ground; Positional accuracy; Course over ground; True heading; True bearing at own position; UTC Seconds. Additionally messages may contain Radio call sign, Vessel Name, Vessel Type; Vessel Dimensions; Vessel Draught, Vessel Destination and vessels estimated time of arrival. Vessels broadcast original position reports at time intervals that vary between 3 seconds and more than 30 seconds, depending on their speed and the type of their AIS transponder. The data streams rate reaches the 17GB/day from tracking more than 65,000 vessels as they travel across the globe.

# 4  Evaluation

## 4.1  Data and pre-Processing

The dataset involved 1.315.474 AIS entries, regarding 2874 unique vessels, including all vessel types (cargo, tankers and passenger vessels, fishing boats and pleasure crafts) as they operate across the Aegean Sea (specifically within the range Latitude from 36.08462 to 39.48708 and Longitude from 24.45557 to 26.58691) for a duration of 30 days, starting on 1/11/2014.

When ranking the various machine learning models, we are using the entirety of our dataset and apply 10-fold cross validation, using the same split in all models. When implementing the real-time data stream analysis service, we need to keep a set of entries in order to be able to evaluate our system. To satisfy this need the data are split based on a timestamp into two datasets at a ratio 3:1; one dataset is used for training purposes and one for evaluation. In particular, the training set contains 986.031 records starting from 2014-11-01 00:00:00 to 2014-11-22 14:59:00. The test set contains 329.443 records starting from 2014-11-22 15:00:00 to 2014-11-30 23:59:00. Some records are automatically filtered out such as those that contain faulty data due to AIS transceiver errors.

Following this, we move on with creating the ground truth. To achieve this we cluster records by vessel and create new fields, i.e. 'Next Longitude', 'Next Latitude' and 'Prediction Interval'. The fields 'Next Longitude' and 'Next Latitude' correspond to the values of the 'Latitude' and 'Longitude' fields from the records with a timestamp that corresponds to the next 'Prediction Interval'.

In this way, we create different datasets with a fixed 'Prediction Interval'. The records of each created set are fewer than the original dataset because in many cases there are no subsequent records with the desired 'Prediction Interval'. This approach is preferred in comparison to create the missing records (using interpolation) because we desired the less possible "noise" in our data.

## 4.2  Machine Learning Models

We constructed two models for each set of different 'Prediction Interval' fields, one for predicting Latitude, and one for Longitude and tested various applicable configurations toying with the model input features (latitude, longitude, speed, course), the prediction algorithm (linear regression, Multi-Layer Perceptron, Random Forest) and the prediction intervals (commonly 2' and 10' but also 20' whenever applicable). Some combinations from the set of these options were not feasible whereas some affected other parameters such as the instances that were considered for training and evaluation of the model. One particular example relates to the amount of subsequent records in a 20' time interval in comparison to the 2' time interval. The results demonstrate the superiority of a 10-node, single hidden layer, multilayer perceptron (MLP) as depicted in Table 1.

The linear regression algorithms (models 1.2-1.20) provide us with an error upper threshold which is then used as a reference value to compare the accuracy of the different models. It is important to note, that during this stage, algorithm performance is of no interest. The main objective of this phase is to identify the model that generates the most accurate results and promote it to implement the data-stream modeling service.

Table 1: Cumulative details for the configurations and results that were tested in order to identify the best performing model, using static data and promote to the real-time, data-stream processing service. The results of the 10-node, single-hidden-layer are highlighted in red. Duration of 10-fold validation is 10 times the training duration.

| | | Configuration | | | | | Results | | | | |
|---|---|---|---|---|---|---|---|---|---|---|---|
| id | Interval | Input Features | Predict | Prediction algorithm | Cross Validation | Total Number of Instances | Correlation coefficient | Mean absolute error | Root mean squared error | Relative absolute error | Root relative squared error | Training duration |
| 1.2 | 2' | [LON, LAT] | LON | Linear Regression | 10 folds | 633803 | 1 | 0.0029 | 0.0038 | 0.52% | 0.60% | 1 '' |
| | | | LAT | | | | 1 | 0.0046 | 0.0053 | 0.71% | 0.66% | 0.55 '' |
| 1.10 | 10' | [LON, LAT] | LON | Linear Regression | 10 folds | 529493 | 0.9996 | 0.0139 | 0.0183 | 2.54% | 2.94% | 0.44 '' |
| | | | LAT | | | | 0.9995 | 0.0225 | 0.0254 | 3.53% | 3.22% | 0.54 '' |
| 1.20 | 20' | [LON, LAT] | LON | Linear Regression | 10 folds | 505478 | 0.9983 | 0.0278 | 0.0365 | 5.11% | 5.89% | 0.50 '' |
| | | | LAT | | | | 0.998 | 0.0443 | 0.0502 | 6.97% | 6.40% | 0.55 '' |
| 2.2 | 2' | [SPEED,LON, LAT,COURSE] | LON | MLP - 10 nodes | 10 folds | 633803 | 1 | 0.0006 | 0.001 | 0.11% | 0.15% | 10.5 ' |
| | | | LAT | | | | 1 | 0.0005 | 0.001 | 0.08% | 0.12% | 10.55 ' |
| 2.10 | 10' | [SPEED,LON, LAT,COURSE] | LON | MLP - 10 nodes | 10 folds | 529493 | 1 | 0.0027 | 0.0049 | 0.50% | 0.78% | 8.7 ' |
| | | | LAT | | | | 1 | 0.002 | 0.0039 | 0.32% | 0.49% | 9.25 ' |
| 2.20 | 20' | [SPEED,LON, LAT,COURSE] | LON | MLP - 10 nodes | 10 folds | 505478 | 0.9998 | 0.0078 | 0.0131 | 1.44% | 2.11% | 8.43 ' |
| | | | LAT | | | | 0.9999 | 0.0071 | 0.0122 | 1.12% | 1.55% | 8.58 ' |
| 3.2 | 2' | [SPEED,LON, LAT,COURSE] | LON | RandomForest - 100 trees | 10 folds | 633803 | 1 | 0.0077 | 0.0097 | 1.40% | 1.55% | 4.09 ' |
| | | | LAT | | | | 1 | 0.0082 | 0.0102 | 1.36% | 1.26% | 4.12 ' |
| 3.10 | 10' | [SPEED,LON, LAT,COURSE] | LON | RandomForest - 100 trees | 10 folds | 529493 | 0.9998 | 0.0087 | 0.0119 | 1.60% | 1.91% | 3.38 ' |
| | | | LAT | | | | 0.9999 | 0.0095 | 0.0124 | 1.49% | 1.57% | 3.42 ' |
| 4.2 | 2' | [SPEED,LON, LAT,COURSE] | LON | MLP - 3 nodes | 10 folds | 633803 | 1 | 0.0033 | 0.0043 | 0.59% | 0.68% | 6.12 ' |
| | | | LAT | | | | 1 | 0.0028 | 0.0036 | 0.39% | 0.59% | 6.18 ' |
| 5.2 | 2' | [SPEED,LON, LAT,COURSE] | LON | MLP - 2Layers - 10,10 nodes | 10 folds | 633803 | 1 | 0.0018 | 0.0028 | 0.34% | 0.45% | 31.2 ' |
| | | | LAT | | | | 1 | 0.0012 | 0.0022 | 0.27% | 0.38% | 32.4 ' |

Tests with different MLP configurations (less nodes, more layers) were performed however they resulted in under-fitting or over-fitting.

SVM models were not used because the complexity of their training is depending on the dataset size. In fact, it is $O(max(n,d) * min(n,d)^2)$, where $n$ is the number of the training documents and $d$ is the number of the input features. Based on this, in the studied case, the time that is required for the training of such a model is prohibitive. Although there are treatments for this problem, such as random sampling (e.g. [29]), however it was decided to use models that can make use of the complete set of data, thus ensuring that no large errors will be propagated to the next step, i.e. using the model for real-time predictions.

### 4.3  Evaluation of the real-time data-stream analysis service

Having achieved most accurate results with the MLP configuration model, we implemented a service that would allow the near real- and fixed-time prediction using data stream data. The service was evaluated against the same dataset which was split to 4 evaluation sets for each of the desired models (4', 10', 20' and 30'). This implies that the MLP model that qualified from the previous step was actually retrained for prediction intervals equal to 4', 10', 20' and 30'. The training took place with the 75% of the original dataset, whereas the rest 25% was used for the evaluation of the online service. An overview of this service in the form of an activity diagram is presented in Figure 2.

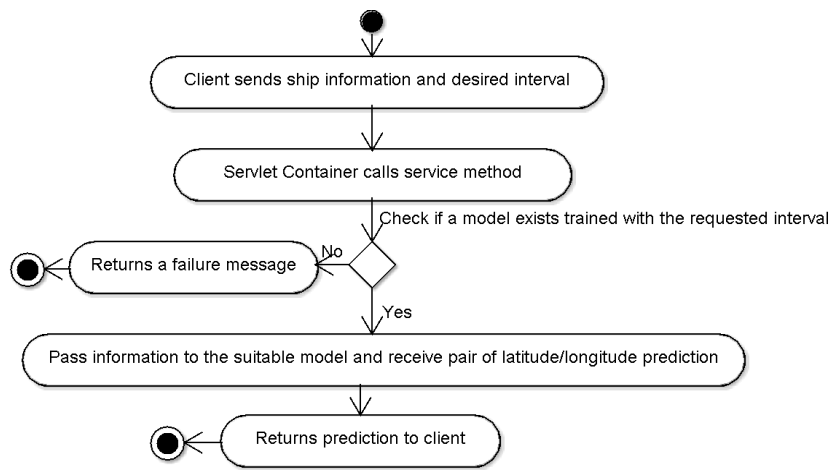

Figure 2: Activity diagram of the predictive service that allows real-time response in future location queries

To evaluate the given models prediction accuracy we calculated the haversive distance. Haversive distance, $d = 2 * r * \arcsin(\sin^2(\Delta\text{lat}/2) + \cos(\text{lat1}).\cos(\text{lat2}).\sin^2(\frac{\Delta\text{long}}{2}))$, gives great-circle distances between two points on a sphere from their longitudes and latitudes, and is a safe approximation for the shortest distance of involving points over the earth's surface.

The dataset allowed for 512.308 predictions, i.e. there were 512.308 records that could be used for testing 4', 10', 20' and 30' time interval predictions cumulatively. These records are attributed almost equally to each of the 4 models (4', 10', 20' and 30').

The server response time (disregarding network latency) for all the predictions was equal to 3706.031106 milliseconds in a Glassfish server configuration running in a 2Gbs JVM. This implies an average 0.00723399ms for each prediction which is by far less than the 0.1 second threshold that enables users feeling that they are freely navigating the command space without having to unduly wait for the computer, even with a network delay in the order of magnitude of 200-300ms [30].

The same experiment was conducted using the simplest kinematic equations, the simplicity of which would intuitively result in a reduced performance (small, fixed response times and zero memory) and comparable accuracy. For calculating the distance, a simple haversive formula is used, that calculates the new location, given the speed, course, latitude and longitude. The acceleration was of no use in this case, as it is not known if the vessel is increasing or decreasing its speed for the interval in question.

Table 2 presents a comparison of the accuracy achieved using the MLP and Kinematic equations models for each prediction interval. It is noticeable how the MLP model surpasses the results of the kinematic equation by a 3.5-5.5x factor.

Table 2: Comparison of the service accuracy when using the MLP model Vs the Kinematic equations

|  |  |  | Mean Haversine Distance (km) | |
|---|---|---|---|---|
| Prediction interval | #Training Instances | #Predictions | MLP model | Kinematic eq. |
| 4' | 434.606 | 135.074 | 0,152521996 | 0,889766298 |
| 10' | 406.976 | 126.888 | 0,652139009 | 2,186280949 |
| 20' | 388.889 | 120.914 | 0,982791041 | 4,256049046 |
| 30' | 435.312 | 129.432 | 1,721464623 | 6,477029597 |

Furthermore, to showcase the suitability of non-time series, machine learning algorithms, in our case, we implemented the proposed model of [6], i.e. a fixed time window stream-processing model with MLP as the learning method.

The model transforms the input, in sets of 10 consecutive vessel positions and predicts the next position. It is evident that due to this requirement, our training (and test) sets will shrink even more. Table 3, presents the instances used for training and testing, as well as the mean haversine distance for each prediction interval.

Table 3: MLP Times Series model metrics

|  |  |  | Mean Haversine Distance (km) |
|---|---|---|---|
| Prediction interval | #Training Instances | #Predictions | MLP Time Series model |
| 4' | 248.682 | 78.059 | 1,314304936 |
| 10' | 229.650 | 72.779 | 1,895986345 |
| 20' | 203.333 | 63.103 | 2,10234085 |
| 30' | 326.484 | 73.545 | 4,612983917 |

Even though a direct comparison between the two approaches cannot be made, since they use different number of training and test instances, we can still derive some interesting insights:

- The time-series approach suffers from the cold start problem; predictions cannot be made for a vessel, until there have been adequate positions recorded in the server. When there is a gap in AIS messages for a particular vessel, either a middle position has to be approximated, or a new series of positions calculated.
- The time-series model will in general require more data in order to manifest comparable performance to the MLP model.
- The time-series model needs more memory and computational power in order to preserve and update the routes of different vessels.

# 5 Conclusions

This work focuses on modeling the trajectory of vessels by analyzing real-time, surveillance, spatiotemporal time series. The model makes no assumptions and categorizations between the vessel characteristics nor it takes into consideration the application semantics, but rather it attempts to process the data stream as it arrives, assuming that each new record contributes equally to a single trajectory forecasting model. This constraint is posed in order to ensure the replicability and scalability properties of the model, to any moving object trajectory prediction model.

Unlike previous approaches in which special implementations of machine learning algorithms were considered, in this work we use a pre-trained model based on traditional multi-scan machine learning algorithms. We demonstrate that the model can assist in responding in real-time queries achieving better accuracy than simple kinematic equations. The model is in fact an "emulated" Kalman filter [31] using machine learning techniques. However, the objective of this work revolves around the issue of performance Vs accuracy tradeoffs therefore the proposed solution is compulsory to be tested disregarding the application domain.

The major contribution of the current work is that it is demonstrated that it is not necessary to resolve to solutions that are tailored to the nature of the big data which in many cases (if not all) sacrifice accuracy in the altar of performance. It is shown that by using traditional machine learning configurations which typically achieve a high gain in terms of accuracy we can implement services performing equally well under real-time constraints. This is proved in a specific application domain (i.e. trajectory prediction) however intuition shows that it is a generalizable solution. It seems that any domain in which a well-trained model can be used for future references without the need for re-training meets these requirements. A final remark regarding this issue is the importance of picking the model configuration that in some cases may be prohibiting for achieving the application goals. For instance, we did not use SVM, as the computation complexity that it would add for the volume of our training set would be prohibitive in meeting the real-time needs.